\documentclass{article}
\usepackage{ijcai21}

\usepackage[round]{natbib}

\usepackage{times}

\usepackage{soul}
\usepackage{url}
\usepackage[hidelinks]{hyperref}
\usepackage[utf8]{inputenc}
\usepackage[small]{caption}
\usepackage{graphicx}
\usepackage{amsmath}
\usepackage{booktabs}
\usepackage{subfigure}

\usepackage{amsfonts}       
\usepackage{amssymb}   
\usepackage{multirow, color}
 \usepackage{algorithm}

\DeclareMathOperator*{\argmax}{arg\,max}

\newtheorem{lemma}{Lemma}

\newtheorem{theorem}{Theorem}
\newtheorem{coro}{Corollary}
\newtheorem{remark}{Remark}

\title{Periodic-GP: Learning Periodic World with Gaussian Process Bandits}

\author{
Hengrui Cai$^1$\footnote{Contact Author. Reinforcement Learning for Intelligent Transportation Systems (RL4ITS) Workshop in the 30th International Joint Conference on Artificial Intelligence (IJCAI-21), Montreal-themed Virtual Reality, Canada, 2021. Copyright 2021 by the author(s).}\and
Zhihao Cen$^2$\and
Ling Leng$^3$\And
Rui Song$^1$
\affiliations
$^1$North Carolina State University 
$^2$INRIA Saclay
$^3$Coupang
\emails
hcai5@ncsu.edu,
zhihao.cen@inria.fr,
Darkfairy.ling@gmail.com,
rsong@ncsu.edu
}

\begin{document}

\maketitle

\begin{abstract}
We consider the sequential decision optimization on the periodic environment, that occurs in a wide variety of real-world applications when the data involves seasonality, such as the daily demand of drivers in ride-sharing and dynamic traffic patterns in transportation. 
In this work, we focus on learning the stochastic periodic world by leveraging this seasonal law. 
To deal with the general action space, we use the bandit based on Gaussian process (GP) as the base model due to its flexibility and generality, and propose the Periodic-GP method with a temporal periodic kernel based on the upper confidence bound. 
Theoretically, we provide a new regret bound of the proposed method, by explicitly characterizing the periodic kernel in the periodic stationary model. Empirically,
the proposed algorithm significantly outperforms the existing methods in both synthetic data experiments and a real data application on Madrid traffic pollution.
\end{abstract}

\section{Introduction} \label{sec:1}
Sequential decision-making is the key component of modern artificial intelligence that considers the dynamics of the real world, and plays a vital role in a wide variety of applications \citep[see e.g.,][]{washburn2008application,gittins2011multi,turvey2017optimal}. By maintaining the exploration-and-exploitation trade-off based on historical information \citep{sutton2018reinforcement}, 
the bandit algorithms thus are popular on dynamic decision optimization \citep{auer2002using,abbasi2011improved}. The goal is to develop an optimal policy over time that maximizes the cumulative reward (or minimizes the cumulative regret) of interest. Most existing bandit algorithms assume a static yet unknown reward mapping function \citep[see the reviews in][]{lattimore2020bandit} that rarely holds in reality as the environment would change over time. For instance, the daily demand on taxi drivers (i.e., the reward function) varies at different times and locations. Typically, it may reach peaks during the morning and evening rush hours around downtown, due to the heavy traffic and commuting. Making use of such seasonal law in ride-sharing could facilitate the repositioning of drivers and further save resources. 

\begin{figure}
\centering
\begin{subfigure}{}
  \centering
\includegraphics[width=0.5\textwidth]{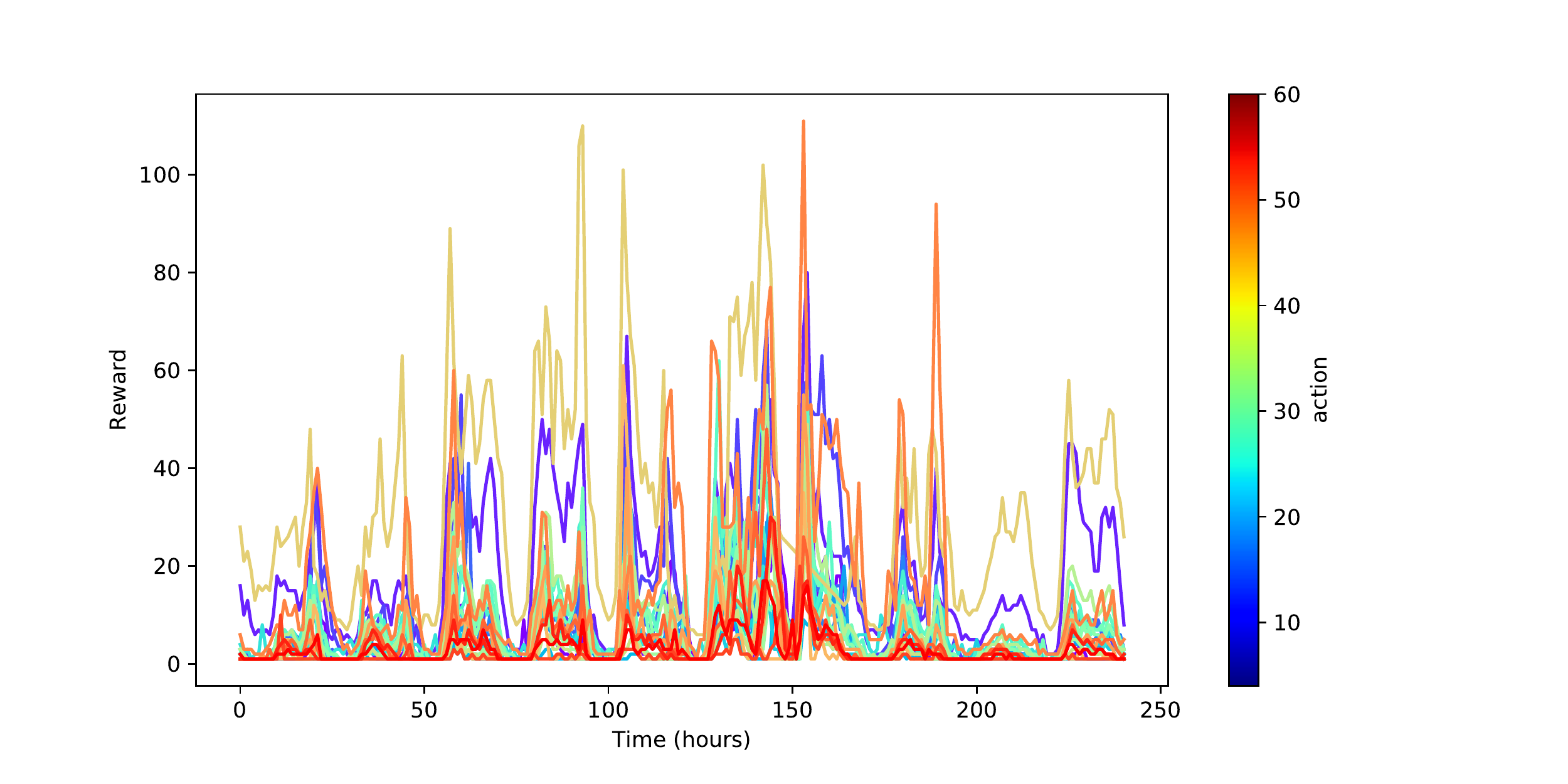}
\end{subfigure}
\caption{Illustration of the reward function (NO,  in $\mu g /m^3$) under different actions (sensor locations) over time (one-hour intervals) in Madrid traffic pollution dataset.} \label{eg_data}
\end{figure}

This periodic phenomenon can be frequently observed in many real-world examples when the data involves seasonality such as the dynamic traffic patterns in transportation. The dataset of Madrid traffic pollution records the traffic condition (i.e., the reward function) from different sensors across Madrid over time. Here, the traffic condition is quantified by the nitric oxide level (NO, measured in $\mu g /m^3$), which is a highly corrosive gas generated by motor vehicles and fuel burning processes. By tracking the NO under different sensor locations (i.e., the actions) over time (as illustrated in Figure \ref{eg_data}), a rapidly changing environment can be observed with strong seasonality of one day period.  
We encode the above environment that periodically repeats with some non-stationary reward functions as `periodic stationary' or  `cyclostationary' \citep{gardner2006cyclostationarity}.  
 Though with a certain known period, the reward function in the periodic stationary environment is usually sophisticated and can change frequently and rapidly within a short time interval. 
Here, the goal is to identify the location at every time step that has the heaviest traffic such that drivers can avoid this traffic jam.

In this work, we develop a new bandit framework to learn the stochastic periodic world by leveraging this seasonal law. 
To deal with the general action space, we use the bandit based on the Gaussian process (GP) as the base model due to its flexibility and generality. Then, we propose the Periodic-GP method with a temporal periodic kernel to address the continuous action with the context of the environment (time step), based on the upper confidence bound (UCB), namely Periodic-GP-UCB. 

Our contributions can be summarized as follows.
Theoretically, we provide the regret bound of our proposed Periodic-GP-UCB, by explicitly characterizing the periodic kernel in the periodic stationary model. Specifically, with a correctly specified circle and the squared exponential kernel, the Periodic-GP-UCB can achieve a regret bound as  $\mathcal{O}\{\sqrt{\tau T(\log T/\tau)^{d+2}}\}$, where $d$ is the dimension of the action space and $\tau$ is the fixed circle parameter in the periodic stationary model of the reward function.
Empirically,
the proposed algorithm significantly outperforms the existing state-of-the-art methods in both synthetic data experiments and a real data application on the Madrid traffic pollution dataset.


 
  \section{Framework}
 \subsection{Problem Statement} 

 Consider to sequentially optimize an unknown reward function $f: \mathcal{A} \times \mathcal{T} \to \mathbb{R}$ over an (potentially infinite) action space $\mathcal{A}$ of $d$ dimensions with an infinite time space $\mathcal{T} = \{1, 2, \cdots\}$. Here, we assume that the time difference between any two consecutive time steps are the same. At each time step $t$, we choose an action $\boldsymbol{a}_t \in \mathcal{A}$ and receive an immediate reward 
 $$y_t =f(\boldsymbol{a}_t,t) +  \epsilon_t,
 $$
  where $\epsilon_t$ is zero mean random noise (independent across the rounds). Let $\boldsymbol{a}^*_t=   \argmax_{a \in \mathcal{A}} f(\boldsymbol{a},t)$ denote a maximizer of $f$ at time step $t$, with $f(\boldsymbol{a}^*_t,t)$ as the best possible mean-reward at time $t$. Then the instantaneous regret incurred at time $t$ is 
  $
  r_t =f(\boldsymbol{a}^*_t,t) -f(\boldsymbol{a}_t,t).$
   The goal is to choose a sequence of actions that minimizes the cumulative regret up to time $T$: 
  $$R_T= \sum_{t=1}^T r_t.$$

\subsection{Periodic Stationary Model} 
Inspired by the Madrid traffic pollution dataset illustrated in Figure \ref{eg_data}, we consider a periodic stationary model to characterize the non-stationary reward function $f$. Specifically, we are interested in the following environment that periodically repeats with the same non-stationary reward function under a fixed known circle $\tau$:
\begin{equation}\label{cyclostationary}
    f(\boldsymbol{a},t) \equiv f\{\boldsymbol{a},g_\tau(t)\}, \quad \forall \boldsymbol{a} \in \mathcal{A}, t =1, 2, \cdots,
\end{equation} 
where $g_\tau(t)$ is a periodic function of time $t$ which satisfies $g_\tau(t) \equiv g_\tau(k\tau + t), \forall k\in \mathbb{N}^+$.


\begin{remark}
Model \ref{cyclostationary} is known as the weak-sense / wide-sense periodic process or cyclostationary process in the time series with independent noise \citep{gardner2006cyclostationarity}. This model assumption is typically valid when the data involves seasonality as discussed in the introduction. A relaxed model assumption is almost-cyclostationary in the wide sense \citep{gardner1986introduction} if $g_\tau$ is an almost periodic function of the time with frequencies not depending on $\tau$.
\end{remark}

\subsection{Related Works}

There are numerous methods proposed for bandits problem with general action space \citep{kleinberg2005nearly, wang2009algorithms, carpentier2015simple}. However, the cited above heavily rely on the prior knowledge of the arm reservoir distribution or continuity of reward function. To alleviate the assumption of the reward function, \cite{srinivas2009gaussian} considered the framework of Gaussian processes (GPs) optimization with kernel on the action space, known as GP-UCB. \cite{krause2011contextual} further proposed the contextual bandits based on the GP-UCB, known as C-GP-UCB. Whereas, all these methods assume a static reward function over time. 

A few bandit algorithms have been proposed to deal with the non-stationary reward \citep[see e.g.,][]{besbes2014stochastic,luo2017efficient,wu2018learning,chen2019new,besbes2019optimal,russac2019weighted,auer2019achieving,cheung2019hedging,trovo2020sliding}. However, none of these methods could solve the periodic environment properly. Specifically, all the cited above rely on a piecewise stationary assumption that the reward function changes arbitrarily at an arbitrary time but remains constant between any two consecutive change points, suffering from three limitations. First, the validity of these methods requires each stationary period is sufficiently long and the difference between two consecutive stationary periods is significantly different. This assumption can be easily violated when the environment changes rapidly or smoothly, such as the example of the Madrid traffic pollution dataset. Second, due to the typically short stationary period in their setting, most of these methods fail to handle the infinitely many or continuous action space that is of great interest in reality. Third, the cited methods only change their belief of the environment with enough evidence, i.e. `action delay', and thus end up with an inefficient regret bound. 

To the best of our knowledge, there are very limited methods designed for the stochastic periodic reward with continuous action yet. Recently, \cite{chen2020learning} studied the multi-armed bandit for unknown periodic reward, though with discrete arms, via Fourier analysis. Another closely related work proposed in \cite{bogunovic2016time} used the GP to address the time-varying environment and the continuous action space, by characterizing the reward function with an exponential kernel as a decay weight. Their proposed R-GP-UCB and TV-GP-UCB methods split the time block into smaller blocks of size, within which the overall variation is assumed to be small. This assumption is fairly close to the piece-wise stationary assumption, and thus fails to handle the periodic stationary environment as discussed previously. We show theoretically that the proposed Periodic-GP-UCB has much improvement beyond these methods with a tighter regret bound (see Section \ref{bound}) in the periodic world. As illustrated in both synthetic and real data analysis (see Section \ref{real_data}), our proposed method significantly outperforms existing state-of-the-art GP-based methods on learning the periodic environment.
 
 
\section{Method}
 We next detail the proposed Periodic-GP method designed for the periodic stationary model in Model \ref{cyclostationary}, which consists of two steps: 1. learn the reward function through the GP with a periodic kernel over time dimension; 2. select the best action based on the upper confidence bound (UCB).
  
\subsection{Reward Learning Algorithm}

Let the action-time space as $\mathcal{S} \equiv \mathcal{A}\times \mathcal{T}$, where each $\boldsymbol{s}=(\boldsymbol{a},t) \in \mathcal{S}$ presents an action-time pair. In this paper, we assume the unknown reward function $f$ is a sample from a Gaussian process (GP) over the action-time space $\mathcal{S}$ as the prior distribution, denoted by $\mathcal{G}\mathcal{P}_\mathcal{S}\{\mu,k\}$ with its mean function $\mu(\boldsymbol{s}) = E[f(\boldsymbol{s})]$ and covariance (or kernel) function $k(\cdot,\cdot): \mathcal{S} \times \mathcal{S} \to \mathbb{R}$ such that $k(\boldsymbol{s},\boldsymbol{s}') = E[\{f(\boldsymbol{s})-\mu(\boldsymbol{s})\}\{f(\boldsymbol{s}')-\mu(\boldsymbol{s}')\}]$. 
Without loss of generality, we assume that $\mu \equiv 0$ with a bounded variance by restricting $k(\boldsymbol{s},\boldsymbol{s}') \leq 1$, for all $\boldsymbol{s},\boldsymbol{s}' \in \mathcal{S}$.  

To characterize the reward function $f$ that satisfies the periodic stationary model in \ref{cyclostationary} on the action-time space $\mathcal{S}$, we consider decomposing the covariance function $k$ into the corresponding covariance functions on actions and times, respectively. More specifically, suppose the periodic function $g_\tau(t)$ can be presented by a periodic kernel \citep{rasmussen2003gaussian} on time with the circle $\tau$:
\begin{equation}\label{kernel_time}
k_\tau(t, t') = \exp\left[-\frac{2}{l^2}\sin^2\Big\{\frac{\pi|t-t'|}{\tau}\Big\}\right],
\end{equation}
where $l$ is the length scale, and $\tau$ is the pre-specified fixed circle. Here, we assume the cycle or period of the periodic environment is known or detectable based on historical information. This assumption is typically valid in a wide variety of examples, such as daily and weekly seasonality for advertising or traffic \citep{karlsson2018control,he2019identification}, and monthly and yearly seasonality for climate or tourism \citep{amelung2007implications,hanninen2011tree}. We include more discussion on circle estimation and model misspecification in Section \ref{sec:7} and sensitivity analysis in the supplementary article when the circle is misspecified.

With a kernel on the action space as $k_\mathcal{A}(\cdot,\cdot): \mathcal{A} \to \mathbb{R}$, we define the kernel function over $\mathcal{S}$ as:
\begin{equation}\label{kernel}
k(\boldsymbol{s}, \boldsymbol{s}') =k\{(\boldsymbol{a},t), (\boldsymbol{a}',t')\}= k_\mathcal{A}(\boldsymbol{a},\boldsymbol{a}') \times k_\tau(t, t').
\end{equation}

Given the sequence of reward received $\boldsymbol{y}_{1:T} =[y_1,\cdots,y_T]^\top$ at inputs $\boldsymbol{s}_{1:T} = [\boldsymbol{s}_1,\cdots,\boldsymbol{s}_T]^\top \equiv \{(\boldsymbol{a}_t, t)\}_{1\leq t \leq T}$ with the noise variables $\epsilon_t$ are drawn independently across $t$ from $N (0,\sigma^2)$, the posterior distribution over $f$ is still a GP with mean $\boldsymbol{\mu}_T(\cdot)$, and covariance $\boldsymbol{\sigma}^2_T(\cdot,\cdot)$:
\begin{equation}\label{post_GP}
\begin{split}
    \boldsymbol{\mu}_T(\boldsymbol{s}) := &\boldsymbol{k}_{1:T}(\boldsymbol{s})^\top  (\boldsymbol{K}_T + \sigma^2 \boldsymbol{I}_T)^{-1} \boldsymbol{y}_{1:T}\\ 
    \boldsymbol{\sigma}^2_T(\boldsymbol{s}, \boldsymbol{s}')  :=&k(\boldsymbol{s}, \boldsymbol{s}') -  \boldsymbol{k}_{1:T}(\boldsymbol{s})^\top (\boldsymbol{K}_T + \sigma^2 \boldsymbol{I}_T)^{-1} \boldsymbol{k}_{1:T}(\boldsymbol{s}')^\top,
\end{split}
\end{equation}
where $\boldsymbol{k}_{1:T}(\boldsymbol{s})  = [k(\boldsymbol{s}_1,\boldsymbol{s}),\cdots, k(\boldsymbol{s}_T,\boldsymbol{s})]^\top$, $\boldsymbol{I}_T$ is the $T\times T$ identity matrix, and $\boldsymbol{K}_T = [k(\boldsymbol{s}, \boldsymbol{s}')]_{\boldsymbol{s},\boldsymbol{s}'\in \boldsymbol{s}_{1:T}}$ is the $T\times T$ kernel matrix at time $T$ with element at $i$-th row and $j$-th column as $k(\boldsymbol{s}_i, \boldsymbol{s}_j)$.


\subsection{Action Selection Strategy}
We next introduce the common action selection strategy used in the GP bandits \citep{srinivas2009gaussian,krause2011contextual,bogunovic2016time}, and extend it into the periodic stationary model.
 
Given collected observations up to times step $t-1$, we can update the posterior mean and variance of the underlying reward function $f$ based on Equation \ref{post_GP}, denoted as $\boldsymbol{\mu}_{t-1}$ and $\boldsymbol{\sigma}^2_{t-1}$. For the next time step $t$, by noting the time information is given, the second element of the action-time pair is always fixed as $t$. Therefore, we have the mean $\boldsymbol{\mu}_{t-1}\{(\boldsymbol{a},t)\}$ and variance $\boldsymbol{\sigma}^2_{t-1}\{(\boldsymbol{a},t)\}$ are pure functions over the action space. Then, we select the action that maximizes the upper confidence bound (UCB) by:
\begin{eqnarray} \label{ucb}
\boldsymbol{a}_{t} = \argmax_{\boldsymbol{a}\in \mathcal{A}} \Big[\boldsymbol{\mu}_{t-1}\{(\boldsymbol{a},t)\} +\beta_t^{-\frac{1}{2}} \boldsymbol{\sigma}_{t-1} \{(\boldsymbol{a},t)\} \Big], 
\end{eqnarray}
where $\beta_t$ is the updating function over time. Note that the above UCB is constructed based on the current information of time step $t$, while we only search over the action space and find the best action for time $t$.

Here, for $\mathcal{A}$ with finite size $|\mathcal{A}|$, we choose 
\begin{eqnarray} \label{beta_dis}
\beta_t= 2 \log(|\mathcal{A}| t^2 \pi^2/ 6\delta),
\end{eqnarray}
where $\delta\in(0,1)$. Besides, for $\mathcal{A} \subset [-v/2, v/2]^d$ is compact and convex with $d \in \mathbb{N}$ and a constant $v > 0$, we update \begin{eqnarray} \label{beta_con}
\begin{split}
\beta_t = & 2 log\{  \pi^2 t^2 /(3\delta)\} \\
&+   2d \log\{c_2  vd t^2\sqrt{\log(c_1d \pi^2 t^2/(3\delta)}\},
\end{split}
\end{eqnarray}
for some constants $c_1,c_2 > 0$. 

We name the above algorithm as Periodic-GP-UCB, with a pseudo-code provided in the following Algorithm 1. The associated regret bound of the proposed Periodic-GP-UCB under different  choices of $\beta_t$ is given in Section \ref{bound}. 

 
\begin{algorithm}\label{C-GP-UCB} 
\caption{Periodic GP-UCB} 
 \textbf{Input}: a pre-specified $\tau$.\\
$1. $  \textbf{While} $t<=T$: \\
$2. \quad$  Set search domain $\boldsymbol{s}=(\cdot,t)$;\\
$3. \quad$ update $\beta_t$ by Equation \ref{beta_dis} for discrete actions or by Equation \ref{beta_con} for continuous actions;\\
$4. \quad$  $\boldsymbol{a}_t \gets \underset{\boldsymbol{a}\in \mathcal{A}}{\argmax} \Big[\boldsymbol{\mu}_{t-1}\{(\boldsymbol{a},t)\} +\beta_t^{-\frac{1}{2}} \boldsymbol{\sigma}_{t-1} \{(\boldsymbol{a},t)\} \Big]$;\\
$5. \quad$  Observe $y_t$;\\
$6. \quad$  Update $\boldsymbol{\mu}_t$ and $\boldsymbol{\sigma}_t$ based on Equation \ref{post_GP};\\

\end{algorithm}

\section{Regret Analysis}\label{bound}
 
We next give the regret bound on the proposed Periodic-GP-UCB. 
Following the same logic in \cite{srinivas2009gaussian}, we show that the regret is bounded by an intuitive information-theoretic quantity, which quantifies the mutual information between the observed action-time pairs and the estimated function $f$ under the periodic stationary model. In addition, we also elaborate on the connection and improvement of the proposed method to the existing GP-based methods. 

\subsection{Preliminary Definitions}
We start with the definition of the maximum mutual information gain under a stationary $f_0$ up to time step $T$ for the action space $\mathcal{A}$ as:
\begin{eqnarray*}
\gamma^ \mathcal{A}_{T}:=\max_{A\subset \mathcal{A}:|A|=T} I(\boldsymbol{y}_{A};f_0),
\end{eqnarray*}
where $A$ is a subset of  $\mathcal{A}$ with size $T$,  and $\boldsymbol{y}_{A} =\{y(\boldsymbol{a})\}_{\boldsymbol{a}\in \mathcal{A}}$. Here, $I(\boldsymbol{y}_{A};f_0) = H(\boldsymbol{y}_{A}) - H(\boldsymbol{y}_{A}|f_0)$ quantifies the reduction in uncertainty about $f_0$ given $\boldsymbol{y}_{A}$, measured in terms of difference of the entropy. Under the multivariate Gaussian, the entropy can be computed explicitly by $H\{N(\boldsymbol{\mu}, \boldsymbol{\Sigma})\} = {1\over2}\log |2\pi e \boldsymbol{\Sigma}|$, so that $I(\boldsymbol{y}_{A};f_0) = {1\over2}\log | \boldsymbol{I} + \sigma^{-2}\boldsymbol{K}_A |$, where $\boldsymbol{K}_A = [k_\mathcal{A}(\boldsymbol{a}, \boldsymbol{a}')]_{\boldsymbol{a}, \boldsymbol{a}'\in A}$ is the Gram matrix of $k_\mathcal{A}$ evaluated on the set $A \subset \mathcal{A}$.

Under the periodic stationary model, we have observations $\boldsymbol{y}_{S} =\{y(\boldsymbol{s})\}_{\boldsymbol{s}\in \mathcal{S}}$ of the joint action-time pairs $\boldsymbol{s} = (\boldsymbol{a},t)$ for $t=1,2,\cdots, T$ and $\boldsymbol{a}\in A\subset \mathcal{A}$ with size $|A|=T$, and an unknown reward function over the action-time space $f:\mathcal{A} \times \mathcal{T} \to \mathbb{R}$. Here, one should note the information of time is given by the system and cannot be changed, so the second element of $\boldsymbol{s}$ is always fixed by its current time step. We then define the new maximum mutual information gain depends on $\boldsymbol{y}_{S}$ and $f$ as:
\begin{eqnarray*}
\begin{split}
\gamma^ \mathcal{S}_{T}&:=\max_{A\subset \mathcal{A}:|A|=T} I(\boldsymbol{y}_{S};f)  =  \max_{A\subset \mathcal{A}:|A|=T}  {1\over2}\log | \boldsymbol{I} + \sigma^{-2}\boldsymbol{K}_S |,
\end{split}
\end{eqnarray*}
where the kernel matrix $\boldsymbol{K}_S = [k(\boldsymbol{s}, \boldsymbol{s}')]_{\boldsymbol{s}, \boldsymbol{s}'\in S}$ is defined over action-time pairs. Based on Equation \ref{kernel}, we have:
\begin{eqnarray*}
\begin{split}
\boldsymbol{K}_S &= [k(\boldsymbol{s}, \boldsymbol{s}')]_{\boldsymbol{s}, \boldsymbol{s}'\in S}\\
& = [k_\mathcal{A}(\boldsymbol{a},\boldsymbol{a}') \times k_\tau(t, t').
]_{(\boldsymbol{a},t), (\boldsymbol{a}',t')\in S} = \boldsymbol{K}_A \circ \boldsymbol{K}_\tau,
\end{split}
\end{eqnarray*}
where $\circ$ presents the Hadamard product, $\boldsymbol{K}_A = [k_\mathcal{A}(\boldsymbol{a}, \boldsymbol{a}')]_{\boldsymbol{a}, \boldsymbol{a}'\in A}$, and $\boldsymbol{K}_\tau = [k_\tau(t, t') = \exp\left[-\frac{2}{l^2}\sin^2\Big\{\frac{\pi(t-t')}{2\tau}\Big\}\right]]_{t,t'\in\{1,2,\cdots, T\}}$ has rank $\tau$.

\subsection{Regret Bound of Periodic-GP-UCB}

Using the new notion of information gain $\gamma^ \mathcal{S}_{T}$, we lift the results of \cite{srinivas2009gaussian} to the periodic stationary model, and derive the regret bound for the Periodic-GP-UCB. The proof of Theorem \ref{thm1} can be found in the supplementary article.

\begin{theorem}\label{thm1}

Let $\delta\in(0,1)$, and $\tau$ is a fixed constant. Suppose one of the following assumptions holds:

1. $\mathcal{A}$ is finite with size $|\mathcal{A}|$, $f$ is sampled from a known GP prior with known noise variance $\sigma^2$, and 
$$
\beta_t= 2 \log(|\mathcal{A}| t^2 \pi^2/ 6\delta);
$$

2. $\mathcal{A} \subset [-v/2, v/2]^d$ is compact and convex with $d \in \mathbb{N}$ and a constant $v > 0$. Suppose $f$ is sampled from a known
GP prior with known noise variance $\sigma^2$, and satisfies the following high
probability bound on the derivatives of GP sample paths for some constants $c_1,c_2 > 0$ at time $t$ where $\boldsymbol{a}=[a^{(1)},\cdots,a^{(d)}]^\top$:
\begin{eqnarray*}
Pr \Bigg\{\sup_{\boldsymbol{a}\in\mathcal{A}} \Bigg|{\partial f(\boldsymbol{a},t) \over \partial a^{(j)}} \Bigg| \geq L_t \Bigg\} \leq c_1 \exp\{-(L_t/c_2)^2\},\\
\quad j = 1,\cdots, d;
\end{eqnarray*}
and update 
\begin{eqnarray*}
\begin{split}
\beta_t = & 2 log\{  \pi^2 t^2 /(3\delta)\}  +   2d \log\{c_2  vd t^2\sqrt{\log(c_1d \pi^2 t^2/(3\delta)}\}.
\end{split}
\end{eqnarray*}

Under condition 1 or 2, we have the regret bound for the Periodic-GP-UCB as $\mathcal{O}\Big(\sqrt{ T  \beta_T \gamma^ \mathcal{S}_{T} } \Big)$ with probability at least $1-\delta$. Or equivalently, we have:
\begin{eqnarray*}
Pr \Bigg\{ R_T \leq \sqrt{c_3 T  \beta_T\gamma^ \mathcal{S}_{T} }+\pi^2/6 , \quad \forall T \geq 1\Bigg\} \geq 1 -\delta.
\end{eqnarray*}
where $c_3 = 8/ \log(1 + \sigma^{-2})$. 

\end{theorem}

\begin{remark}
Here, Theorem \ref{thm1} is built based on the similar techniques used in \cite{srinivas2009gaussian} with a periodic stationary model setting. The results in Theorem \ref{thm1} show that the cumulative regret of the proposed Periodic-GP-UCB is bounded by the maximum mutual information gain ($\gamma^ \mathcal{S}_{T}$), which is consistent with the existing literature on the GP-based bandits \citep{srinivas2009gaussian,krause2011contextual,bogunovic2016time}. 
The way we treat this $\gamma^ \mathcal{S}_{T}$ is unique and novel, with more details provided in the following section.
 \end{remark}
%
%

\subsection{Characterize $\gamma^ \mathcal{S}_{T}$ in Periodic-GP-UCB}

In this section, we provide details to characterize $\gamma^ \mathcal{S}_{T}$ in the Periodic-GP-UCB. All the proofs can be found in the supplementary article.

Let the circle $\tau$ is a fixed constant. Suppose $f$ is sampled from a known GP prior with known noise variance $\sigma^2$. We first bound the mutual information gain by rearranging the time steps $\{1,\cdots, T\}$ into $\tau$ sets of length $T/\tau$, such that within each set the function $f$ is time-invariant, by the following lemma.

 \begin{lemma}\label{mig}
Assume that $K=T/\tau$ is an integer for the time being. Given the sequence of reward received $\boldsymbol{y}_{1:T} =[y_1,\cdots,y_T]^\top$ at inputs $\boldsymbol{s}_{1:T} = [\boldsymbol{s}_1,\cdots,\boldsymbol{s}_T]^\top \equiv \{(\boldsymbol{a}_t, t)\}_{1\leq t \leq T}$ with the noise variables $\epsilon_t$ are drawn independently across $t$ from $N (0,\sigma^2)$, the overall mutual information gain of $\boldsymbol{y}_{1:T}$ given $\boldsymbol{f}_{1:T}$ is bounded by the summation of the mutual information gain of its subsets in the periodic stationary model:
\begin{eqnarray*}
I(\boldsymbol{y}_{1:T};\boldsymbol{f}_{1:T}) \leq  \sum_{i=1}^\tau I\{\boldsymbol{y}^{(i)};\boldsymbol{f}^{(i)}\},
\end{eqnarray*}
where $\boldsymbol{y}^{(i)} =[y_i, y_{i+\tau},\cdots,y_{i+(K-1)\tau}]^\top$, and $\boldsymbol{f}^{(i)} =\{f(\boldsymbol{a}_j,i)\}_{j=i+(k-1)\tau, k=1,\cdots,K}$.
\end{lemma}

\begin{remark}
The result in Lemma \ref{mig} is essential to characterize $\gamma^ \mathcal{S}_{T}$ in the Periodic-GP-UCB, and is new to the GP-based bandit literature. The key idea is to rearrange the mutual information gain sequences by the periodic feature, based on the information theory. The equality in Lemma \ref{mig} can be achieved if and only if the rewards received are independent.  In other words, if the current reward received  depends on the previous information, we have the mutual information gain under the proposed Periodic-GP is strictly smaller than $\tau$ baseline GP models.
\end{remark}

Next, we bound the $\gamma^ \mathcal{S}_{T}$ in the following theorem, based on the result in Lemma \ref{mig}.
 
\begin{theorem}\label{thm2}
Assume that $K=T/\tau$ is an integer for the time being. Given the sequence of reward received $\boldsymbol{y}_{1:T} =[y_1,\cdots,y_T]^\top$ at inputs $\boldsymbol{s}_{1:T} = [\boldsymbol{s}_1,\cdots,\boldsymbol{s}_T]^\top \equiv \{(\boldsymbol{a}_t, t)\}_{1\leq t \leq T}$ with the noise variables $\epsilon_t$ are drawn independently across $t$, the maximum mutual information $\gamma^ \mathcal{S}_{T}$ in Periodic-GP satisfies
$$
\gamma^ \mathcal{S}_{T} \leq  \sum_{i=1}^\tau  \gamma^ \mathcal{A}_{T/\tau} = \tau  \gamma^ \mathcal{A}_{K}.
$$
 \end{theorem}

\begin{remark}
When $\tau =1$, i.e., the environment is stationary, we have $\gamma^ \mathcal{S}_{T} = \gamma^ \mathcal{A}_{T}$ based on Theorem \ref{thm2}. In other words, the GP-based bandit in \cite{srinivas2009gaussian} can be treated as a special case of the Periodic-GP framework. Similarly, the equality in Theorem \ref{thm2} can only be achieved when the sequential rewards are independent. Thus, the maximum mutual information gain under the Periodic-GP is strictly smaller than $\tau$ GP models, when the current reward depends on the previous information.
\end{remark}

Under Theorem \ref{thm2}, with a correctly specified circle in the periodic stationary model, we can further quantify $\gamma^ \mathcal{S}_{T}$ in the Periodic-GP-UCB under different kernel functions on the action space, based on the results of Theorem 5 in \cite{srinivas2009gaussian}.

\begin{coro}\label{coro1}
Assume the conditions in Theorem \ref{thm2} hold, and the kernel function on the action space (with $d$ dimension) satisfies $k_\mathcal{A}(\boldsymbol{a}, \boldsymbol{a}')\leq 1$. With a correctly specified circle in the periodic stationary model, we have the regret bound of $\gamma^ \mathcal{S}_{T}$ under the following kernel functions as: 

1. For the $d$-dimensional linear kernel 
$k_\mathcal{A}(\boldsymbol{a}, \boldsymbol{a}')=\boldsymbol{a}^\top \boldsymbol{a}',$
 we have
$$\gamma^ \mathcal{S}_{T} = \mathcal{O}\{d\tau(\log T/\tau) \};$$
2. For the squared exponential kernel 
$k_\mathcal{A}(\boldsymbol{a}, \boldsymbol{a}')=\exp\big\{-  || \boldsymbol{a} - \boldsymbol{a}' ||^2 / ({2l^2})\big\},$
with a length scale $l$,  we have 
$$\gamma^ \mathcal{S}_{T} = \mathcal{O}\{\tau(\log T/\tau)^{d+1}\};$$
3. For the mat\'ern kernels 
$
k_\mathcal{A}(\boldsymbol{a}, \boldsymbol{a}')=  \{2^{1-\nu}/\Gamma(\nu)\} \{(\sqrt{2\nu}/l) || \boldsymbol{a} - \boldsymbol{a}' ||\} K_\nu\{(\sqrt{2\nu}/l) || \boldsymbol{a} - \boldsymbol{a}' ||\},$
where $\Gamma$ is the gamma function, $K_\nu$ is the modified Bessel function, and $\nu$ is positive parameter that controls the smoothness of sample paths, we have
$$
\gamma^ \mathcal{S}_{T} = \mathcal{O}\{(T/\tau)^{d(d+1)/\{2\nu+d(d+1)\}}\tau(\log T/\tau) \}.
$$
 \end{coro}

 The above corollary is a direct result of Theorem \ref{thm2} in this paper together with Theorem 5 in \cite{srinivas2009gaussian}. 
Based on Corollary \ref{coro1}, it can be shown that the proposed Periodic-GP-UCB has a similar order of the regret bound as in the GP-UCB  \citep{srinivas2009gaussian}. Specifically, with the squared exponential kernel, the Periodic-GP-UCB can achieve a regret bound as $\mathcal{O}\{\sqrt{\tau T(\log T/\tau)^{d+2}}\}$. In contrast, the regret bounds in the R-GP-UCB and the TV-GP-UCB  \citep{bogunovic2016time} are merely approaching $\mathcal{O}(T)$, due to the large overall variation of a given time-interval under the considered periodic stationary model. Therefore, by making use of the temporal periodic kernel, we obtain a tighter regret bound for the periodic environment. 


 
\section{Experiments}\label{real_data}

In this section, we conduct extensive experiments to compare the proposed Periodic-GP-UCB to the GP-UCB in \cite{srinivas2009gaussian} and the C-GP-UCB in \cite{krause2011contextual} for the stationary environment, as well as the R-GP-UCB and the TV-GP-UCB in \cite{bogunovic2016time} for the time-varying environment. A sensitivity analysis is provided in the supplementary article when the cycle $\tau$ is misspecified, where our proposed method still maintains a good performance under minor assumption violation.

\subsection{Synthetic Data}

\begin{figure} 
\centering
\begin{subfigure} 
  \centering
\includegraphics[width=0.5\textwidth]{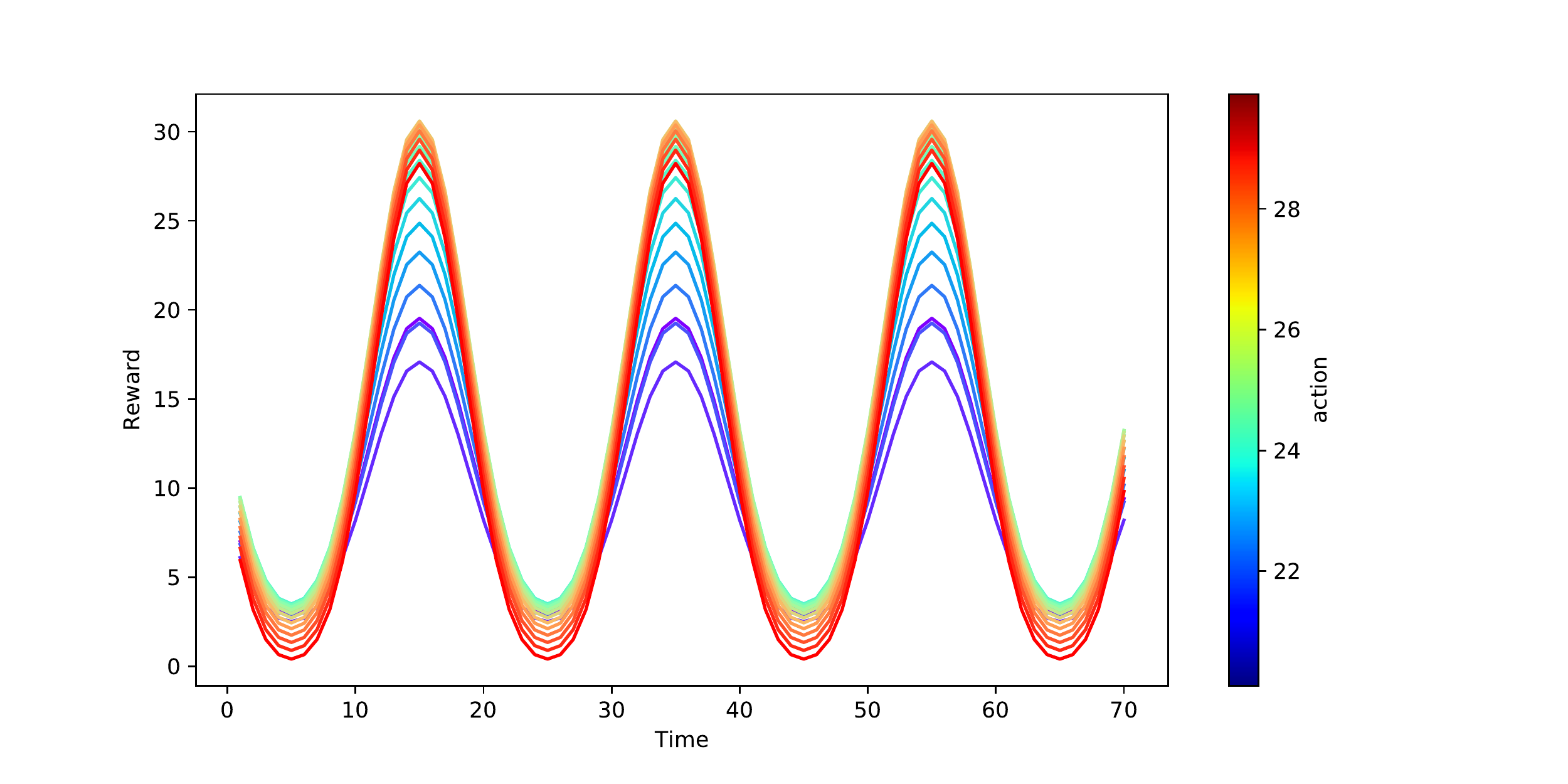} 
\end{subfigure} 
\caption{The reward function under different actions over time for synthetic data.}\label{scen}
\end{figure}

 We consider a one-dimensional continuous action space, and generate our data according to a smoothly changing environment for synthetic datasets, as illustrated in Figure \ref{scen} with the first 70 time steps. The seasonality circle is set to be $\tau = 20$. We consider the squared
exponential kernel for the action space with the length-scale $l = 1$, and the periodic kernel (see Equation \ref{kernel_time}) for the time space with the length-scale as $l = 10$. The sampling noise follows $N(0,1)$.

\begin{figure}[!t]
\centering
\begin{subfigure}{}
  \centering
\includegraphics[width=0.47\textwidth]{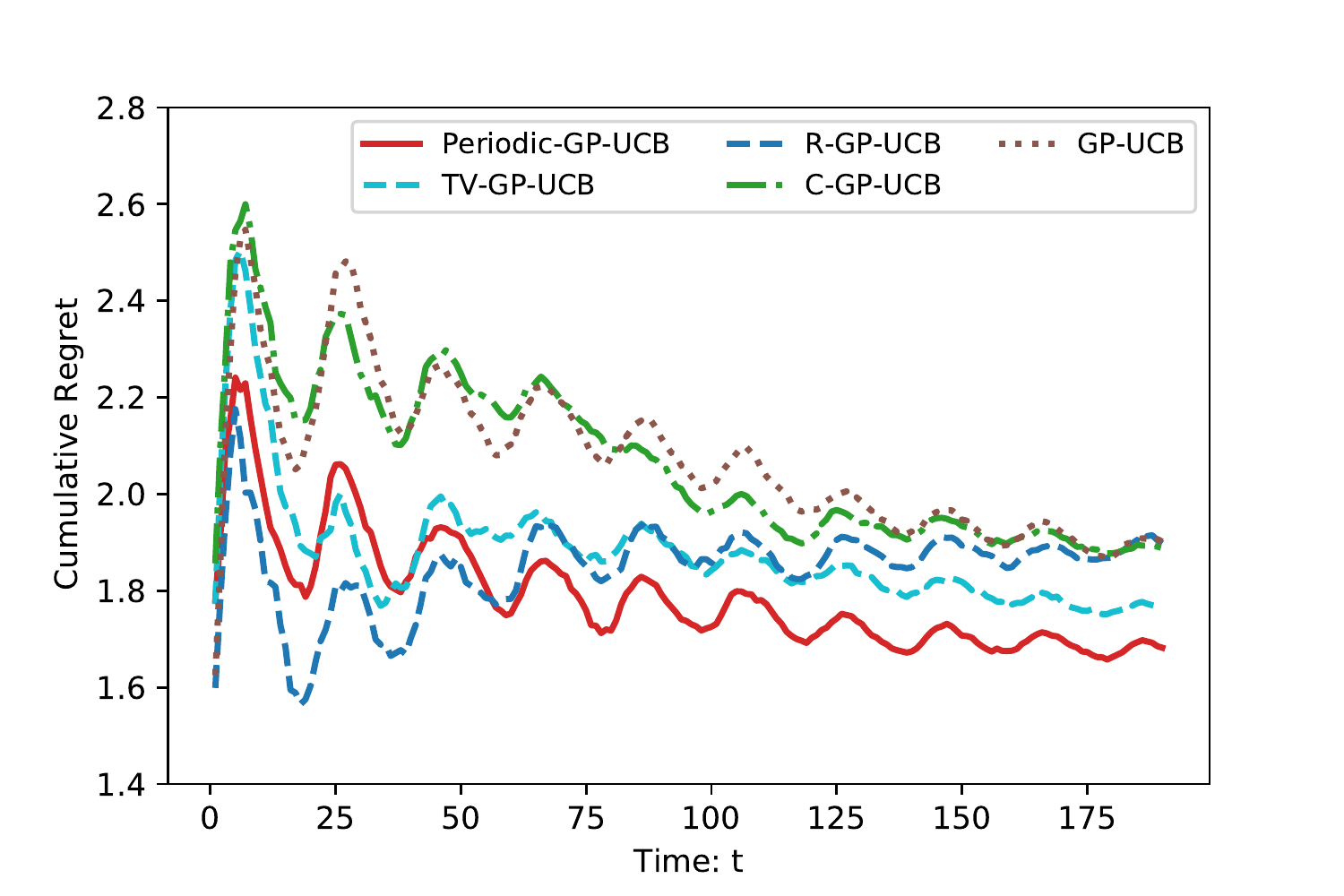}
\end{subfigure} 
\caption{The mean cumulative regret over time under different methods for synthetic data.}\label{simu_res}
\end{figure}

We apply the proposed method for the total time steps as $T=200$ with 100 replications, in comparison to four GP-based bandits. For the C-GP-UCB in \cite{krause2011contextual}, we consider using the time step as the contextual information with the squared
exponential kernel with the same length-scale as $l = 10$. In addition, we set the block size as 15 for the R-GP-UCB as suggested in \cite{bogunovic2016time} and choose the variance bound term $\epsilon$ in the TV-GP-UCB based on the pre-trained model. To be consistent with the existing methods  \citep{srinivas2009gaussian,krause2011contextual,bogunovic2016time}, we also consider $\beta_t=0.8\log(0.4 t)$ for a fair comparison. 
The performance is evaluated by the cumulative regret averaged over 100 replications, as shown in Figure \ref{simu_res}.

It can be observed in Figure \ref{simu_res} that the proposed Periodic-GP-UCB (red solid curves) achieves the best performance among all the methods. Specifically, the R-GP-UCB (cyan dashed curves) and the TV-GP-UCB (blue dashed curves) have relatively better performance than the GP-UCB and the C-GP-UCB.
This observation is due to the smoothly changing environment, i.e., both stationary and piece-wise stationary assumptions are seriously violated here, which implies that the bandits designed for stationary or piece-wise stationary model fail to address the periodic environment, as discussed in the introduction and related works. In addition, by simply adding the time information as the context into the bandit, there is no much improvement of the C-GP-UCB (green dotted and dashed curves) compared to the baseline GP-UCB (brown dotted curves), which indicates that such exploitation in the time space is not efficient. In contrast, by introducing the temporal periodic kernel, our proposed Periodic-GP-UCB could reduce the cumulative regret in the changing environment.

\subsection{Madrid Traffic Pollution Data}

We use the Madrid Traffic Pollution data with 24 sensors collected from the Madrid's City Council Open Data website \footnote{https://datos.madrid.es/portal/site/egob} for analysis. The dataset contains the measurements from March 3rd, 2018 to March 13th, 2018, as a total of 10 days collected at one-hour intervals. Our goal is to identify the location at every time step that has the heaviest traffic. The regret is defined by the difference between the nitric oxide level (NO, measured in $\mu g /m^3$) at the selected location and the maximum NO level at that particular time. The values of the NO pollution under different sensor locations over time are illustrated in the upper panel of Figure \ref{eg_data}, where a reasonable periodic stationary pattern can be observed with one day period, while the environment changes rapidly.

\begin{figure}
\centering
\begin{subfigure}{}
  \centering
\includegraphics[width=0.47\textwidth]{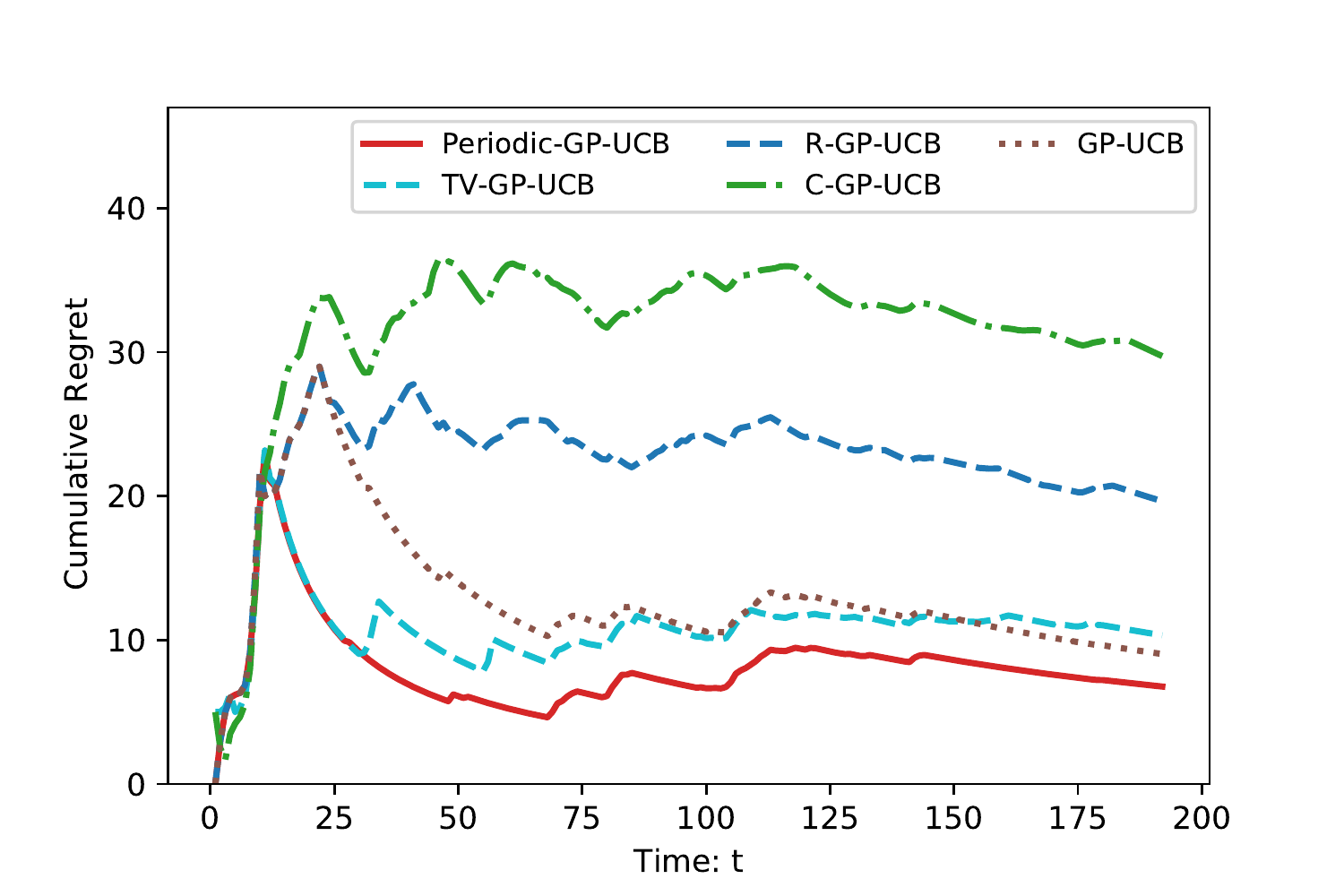} 
\end{subfigure} 
\caption{The cumulative regret under different methods for the Madrid traffic pollution data.} \label{real_res}
\end{figure}

We apply the Periodic-GP-UCB with $\tau =24$ against four GP-based bandits. All the hyper-parameters are set based on the pre-trained model as mentioned in the previous section. Here, we use the best action-reward pairs during the first two days of the dataset as the prior information for training all methods. The algorithms are run on the rest of eight days of the dataset for testing. Similarly, we set $\beta_t=0.8\log(0.4 t)$ with the signal variance $\sigma^2 = 0.5$ in each method. The cumulative regret as reported in the upper panel of Figure \ref{real_res} under different algorithms.

It can be observed that the proposed Periodic-GP-UCB outperforms the existing methods, followed by the GP-UCB that is designed for the stationary environment. In other words, the TV-GP-UCB and the R-GP-UCB fail to capture the rapidly changing environment in the Madrid traffic pollution, and eventually worse than a stationary guess. Moreover, the C-GP-UCB performs the worst since it ignores the periodic stationary pattern of the environment and purely takes the time information as the context. In contrast, by introducing the temporal periodic kernel in our proposed Periodic-GP-UCB, one can reduce the cumulative regret by 13\%, compared to the GP-UCB (the second-best method). 

\section{Discussion} \label{sec:7}
In this paper, we focus on the common pattern of the non-stationary environment, periodic stationary, and develop the Periodic-GP method with a temporal periodic kernel to address the general class of action. 
Theoretically, we show the regret bound of our proposed Periodic-GP-UCB by explicitly characterizing the periodic kernel in the periodic stationary model. This regret bound, to the best of our knowledge, is new to the bandit literature, and is tighter compared to the methods designed for the piece-wise stationary model.
Empirically,
the proposed algorithm significantly outperforms the existing state-of-the-art methods in both synthetic and real datasets.

We conclude our paper by following possible extensions. First, the period $\tau$ of the periodic kernel in Equation \ref{kernel_time} may be misspecified. In practice, we can use the historical data to verify the cycle length or pattern by detecting the seasonality in the observed rewards given a fixed action \citep{walter1975test}. 
Second, 
we can extend our method with contextual information, to develop the dynamic individualized policy that maps each context to the best action to maximize the overall non-stationary reward. Third, 
the current Periodic-GP method can be generalized to address a switching periodic environment, by combining the periodic kernel in Equation \ref{kernel_time} with an extra exponential decay kernel considered in \cite{bogunovic2016time} on the time space.

\bibliography{ijcai21} 
\bibliographystyle{named}
\appendix
 \onecolumn 
\begin{center}
\textbf{\LARGE Supplementary to `Periodic-GP: Learning Periodic World with Gaussian Process Bandits'}
\end{center}

 \section{Proof of Theorem 1}
In this section, we provide details for the proof of Theorem 1, by modifying and generalizing the results of \cite{srinivas2009gaussian} to the periodic stationary model. Let $\delta\in(0,1)$, and the circle $\tau$ is a fixed constant. We first extend the results of the mutual information gain under our proposed model, and then consider three different assumptions on the action space $\mathcal{A}$ or the true underlying $f$, respectively. With an appropriate choice of the updating parameter $\beta_t$ for each assumption, we will show the regret bound for Periodic-GP-UCB is $\mathcal{O}\Big(\sqrt{ T  \beta_T \gamma^ \mathcal{S}_{T} } \Big)$ with probability at least $1-\delta$ in all three cases, where $\gamma^ \mathcal{S}_{T}$ is  the maximum mutual information gain depends on rewards received and $f$ in the periodic stationary model.

\subsection{Results on the Mutual Information Gain}

We start to generalize Lemma 5.3 in \cite{srinivas2009gaussian} to the periodic stationary model. The following result will be used to drive the regret bound for the proposed Periodic-GP-UCB.

\begin{lemma}\label{mig}
Given the sequence of reward received $\boldsymbol{y}_{1:T} =[y_1,\cdots,y_T]^\top$ at inputs $\boldsymbol{s}_{1:T} = [\boldsymbol{s}_1,\cdots,\boldsymbol{s}_T]^\top \equiv \{(\boldsymbol{a}_t, t)\}_{1\leq t \leq T}$ with the noise variables $\epsilon_t$ are drawn independently across $t$ from $N (0,\sigma^2)$, the mutual information gain can be expressed in terms of the predictive variances in the periodic stationary model:
\begin{eqnarray*}
I(\boldsymbol{y}_{1:T};\boldsymbol{f}_{1:T}) = {1\over2}\sum_{t=1}^T \log\Big(1+ \sigma^{-2} \boldsymbol{\sigma}^2_{t-1} \{(\boldsymbol{a}_t,t)\}  \Big),
\end{eqnarray*}
where $\boldsymbol{f}_{1:T} =\{f(\boldsymbol{s}_t)\}$ at the joint action-time pairs $\boldsymbol{s}_t = (\boldsymbol{a}_t,t)$ for $t=1,2,\cdots, T$, and $ \boldsymbol{\sigma}^2_{t-1} \{(\boldsymbol{a}_t,t)\}$ is the posterior variance of pair $(\boldsymbol{a}_t,t)$ at time step $t-1$.
\end{lemma}

\textbf{Proof of Lemma \ref{mig}:} 

By the definition of the mutual information gain, we have $I(\boldsymbol{y}_{1:T};\boldsymbol{f}_{1:T}) = H(\boldsymbol{y}_{1:T}) - H(\boldsymbol{y}_{1:T}|\boldsymbol{f}_{1:T})$. 

The first entropy term can be decomposed by the entropy of $\boldsymbol{y}_{1:(T-1)}$ as
\begin{eqnarray}\label{decom_entropy}
\begin{split}
H(\boldsymbol{y}_{1:T}) &= H\{\boldsymbol{y}_{1:(T-1)}\} + H\{\boldsymbol{y}_{1:T}|\boldsymbol{y}_{1:(T-1)}\}\\
&= H\{\boldsymbol{y}_{1:(T-1)}\} + {1\over2} \log\Big[2\pi e\{ \sigma^{2} +\boldsymbol{\sigma}^2_{T-1} \{(\boldsymbol{a}_T,T)\}  \Big],
\end{split}
\end{eqnarray}
where the second equation comes from the posterior variance under $\boldsymbol{s}_{1:T}$.

On the other hand, since $y_t =f(\boldsymbol{a}_t,t) +  \epsilon_t$ where $\epsilon_t$ is zero mean with variance $\sigma^2$, the second entropy $H(\boldsymbol{y}_{1:T}|\boldsymbol{f}_{1:T}) = {1\over2} \log | 2\pi e \sigma^2 \boldsymbol{I} | = {1\over2} T \log ( 2\pi e \sigma^2 )$ under the normality assumption.

Thus, by replacing two entropy terms in $I(\boldsymbol{y}_{1:T};\boldsymbol{f}_{1:T})$, we have 
\begin{eqnarray}\label{decom_mig}
\begin{split}
I(\boldsymbol{y}_{1:T};\boldsymbol{f}_{1:T}) &= H\{\boldsymbol{y}_{1:(T-1)}\} + {1\over2} \log\Big[2\pi e\{ \sigma^{2} +\boldsymbol{\sigma}^2_{T-1} \{(\boldsymbol{a}_T,T)\}  \Big] -{1\over2} T \log ( 2\pi e \sigma^2 ),\\
& = H\{\boldsymbol{y}_{1:(T-1)}\} + {1\over2} \log\Big(1+ \sigma^{-2} \boldsymbol{\sigma}^2_{T-1} \{(\boldsymbol{a}_T,T)\}  \Big)  -{1\over2} (T-1) \log ( 2\pi e \sigma^2 ).\\
\end{split}
\end{eqnarray}

By decomposing $H\{\boldsymbol{y}_{1:t}\}$ for $t=1,2,\cdots, T-1$ based on equation \ref{decom_entropy}, we can update the result in equation \ref{decom_mig} step by step. Following the mathematical induction, we have
\begin{eqnarray*}
I(\boldsymbol{y}_{1:T};\boldsymbol{f}_{1:T}) = {1\over2}\sum_{t=1}^T \log\Big(1+ \sigma^{-2} \boldsymbol{\sigma}^2_{t-1} \{(\boldsymbol{a}_t,t)\}  \Big),
\end{eqnarray*}
with $H\{ {y}_1\} = {1\over2}  \log ( 2\pi e \sigma^2 )$.  
$\blacksquare$

\subsection{Finite Action Space}\label{discrete}
We first consider the action space $\mathcal{A}$ is finite with size $|\mathcal{A}|$. Suppose $f$ is sampled from a known GP prior with known noise variance $\sigma^2$. With $\beta_t= 2 \log(|\mathcal{A}| t^2 \pi^2/ 6\delta)$, its  regret bound can be proved by following three steps.

\textbf{Step 1 (uniform bound of $f$):}

Given a fixed time step $t$ with the sequence of reward received $\boldsymbol{y}_{1:(t-1)} =[y_1,\cdots,y_{(t-1)}]^\top$, the inputs $\boldsymbol{s}_{1:(t-1)} = [\boldsymbol{s}_1,\cdots,\boldsymbol{s}_{(t-1)}]^\top \equiv \{(\boldsymbol{a}_i, i)\}_{1\leq i \leq t-1}$ are deterministic. Fix an action $\boldsymbol{a}\in \mathcal{A}$, then $f$ at time $t$ follows
\begin{eqnarray*}
f(\boldsymbol{a}, t) \sim N\Big[\boldsymbol{\mu}_{t-1}\{(\boldsymbol{a},t)\} , \boldsymbol{\sigma}^2_{t-1} \{(\boldsymbol{a},t)\} \Big].
\end{eqnarray*}
Thus, the normal distribution leads to
\begin{eqnarray*}
Pr\Big\{|f(\boldsymbol{a}, t) - \boldsymbol{\mu}_{t-1}\{(\boldsymbol{a},t)\} |> \sqrt{\beta_t} \boldsymbol{\sigma}_{t-1} \{(\boldsymbol{a},t)\} \Big\} \leq \exp\{-{1\over2}\beta_t\},
\end{eqnarray*}
where $\beta_t= 2 \log(|\mathcal{A}| t^2 \pi^2/ 6\delta)$.

By union bound over time step $t \in  \mathcal{T}$ and $\boldsymbol{a} \in \mathcal{A}$, we have 
\begin{eqnarray*}
 |f(\boldsymbol{a}, t) - \boldsymbol{\mu}_{t-1}\{(\boldsymbol{a},t)\} |\leq \sqrt{\beta_t} \boldsymbol{\sigma}_{t-1} \{(\boldsymbol{a},t)\} , \quad \forall \boldsymbol{a} \in \mathcal{A}, \quad \forall t \in  \mathcal{T},
 \end{eqnarray*}
holds with probability at least
\begin{eqnarray*}
1 -  \sum_{t=1}^\infty |\mathcal{A}| \exp\{-{1\over2}\beta_t\} = 1 - \sum_{t=1}^\infty |\mathcal{A}| \exp\{-{1\over2}2 \log(|\mathcal{A}| t^2 \pi^2/ 6\delta)\} = 1 - \sum_{t=1}^\infty 6\delta/ (t^2 \pi^2) = 1- \delta. 
 \end{eqnarray*}
Here, the last equation comes from $\sum_{t=1}^\infty t^{-2} = \pi^2 /6$.

\textbf{Step 2 (bound the instantaneous regret $r_t$):}

By definition of the action $\boldsymbol{a}_t$ chosen at time $t$ and the best action $\boldsymbol{a}^*_t$ at time $t$, we have
\begin{eqnarray}\label{t1_a1_s1}
\boldsymbol{\mu}_{t-1}\{(\boldsymbol{a}_t,t)\} + \sqrt{\beta_t} \boldsymbol{\sigma}_{t-1} \{(\boldsymbol{a}_t,t)\} \geq \boldsymbol{\mu}_{t-1}\{(\boldsymbol{a}^*_t,t)\} + \sqrt{\beta_t} \boldsymbol{\sigma}_{t-1} \{(\boldsymbol{a}^*_t,t)\}.
 \end{eqnarray} 
 
 Based on step 1, with probability at least $1- \delta$, we have
\begin{eqnarray*}
\boldsymbol{\mu}_{t-1}\{(\boldsymbol{a},t)\} - \sqrt{\beta_t} \boldsymbol{\sigma}_{t-1} \{(\boldsymbol{a},t) \} \leq f(\boldsymbol{a}, t)\leq \boldsymbol{\mu}_{t-1}\{(\boldsymbol{a},t)\} + \sqrt{\beta_t} \boldsymbol{\sigma}_{t-1} \{(\boldsymbol{a},t)\} , \quad \forall \boldsymbol{a} \in \mathcal{A}, \quad \forall t \in  \mathcal{T}.
 \end{eqnarray*}
Therefore, it is immediate to yield the following results with probability at least $1- \delta$
\begin{eqnarray}\label{t1_a1_s2}
f(\boldsymbol{a}^*_t, t) \leq  \boldsymbol{\mu}_{t-1}\{(\boldsymbol{a}^*_t,t)\} + \sqrt{\beta_t} \boldsymbol{\sigma}_{t-1} \{(\boldsymbol{a}^*_t,t)\}, 
 \text{ and } f(\boldsymbol{a}_t, t) \geq \boldsymbol{\mu}_{t-1}\{(\boldsymbol{a}_t,t)\} - \sqrt{\beta_t} \boldsymbol{\sigma}_{t-1} \{(\boldsymbol{a}_t,t) \}.
\end{eqnarray}
 
Then, with probability at least $1- \delta$, the instantaneous regret incurred at time $t$ can be bound by
\begin{eqnarray}\label{t1_a1_bound_r_t}
\begin{split} 
r_t &=f(\boldsymbol{a}^*_t,t) -f(\boldsymbol{a}_t,t)\\
&\leq  \boldsymbol{\mu}_{t-1}\{(\boldsymbol{a}^*_t,t)\} + \sqrt{\beta_t} \boldsymbol{\sigma}_{t-1} \{(\boldsymbol{a}^*_t,t)\} - [\boldsymbol{\mu}_{t-1}\{(\boldsymbol{a}_t,t)\} - \sqrt{\beta_t} \boldsymbol{\sigma}_{t-1} \{(\boldsymbol{a}_t,t)\}] \\
&\leq \boldsymbol{\mu}_{t-1}\{(\boldsymbol{a}_t,t)\} + \sqrt{\beta_t} \boldsymbol{\sigma}_{t-1} \{(\boldsymbol{a}_t,t)\} - [\boldsymbol{\mu}_{t-1}\{(\boldsymbol{a}_t,t)\} - \sqrt{\beta_t} \boldsymbol{\sigma}_{t-1} \{(\boldsymbol{a}_t,t)\}] = 2 \sqrt{\beta_t} \boldsymbol{\sigma}_{t-1} \{(\boldsymbol{a}_t,t)\},\\
\end{split}
\end{eqnarray}
where the first inequality is the instant result by inequality \ref{t1_a1_s2}, and the second inequality is from inequality \ref{t1_a1_s1}.
 
 \textbf{Step 3 (bound the cumulative regret $R_T$):}
 
Based on Cauchy-Schwarz inequality, the cumulative regret up to time $T$ is bounded by
\begin{eqnarray*} 
R_T= \sum_{t=1}^T r_t \leq \sqrt{T \sum_{t=1}^T r_t^2},
\end{eqnarray*}     
with probability at least $1- \delta$.

Therefore, from the result in \ref{t1_a1_bound_r_t}, we have
\begin{eqnarray*} 
R_T \leq \sqrt{T \sum_{t=1}^T 4  {\beta_t} \boldsymbol{\sigma}^2_{t-1}\{(\boldsymbol{a}_t,t)\}} \leq \sqrt{ 4 T {\beta_T}\sigma^{2} \sum_{t=1}^T \sigma^{-2}\boldsymbol{\sigma}^2_{t-1}\{(\boldsymbol{a}_t,t)\}},
\end{eqnarray*}  
where $\beta_T =2 \log(|\mathcal{A}| T^2 \pi^2/ 6\delta) \geq 2 \log(|\mathcal{A}| t^2 \pi^2/ 6\delta) = \beta_t$ for $t\in \mathcal{T}$.

Denote $w(x) = x/\log(1+ x) $, which is an increasing function of $x > 0$. Let $c_4 =w(\sigma^{-2}) =  \sigma^{-2}/\log(1+ \sigma^{-2}) \geq 1$. Since $\sigma^{-2}\boldsymbol{\sigma}^2_{t-1}\{(\boldsymbol{a}_t,t)\} = \sigma^{-2}k\{(\boldsymbol{a}_t,t), (\boldsymbol{a}_t,t)\} \leq \sigma^{-2}$, we have $w[\sigma^{-2}\boldsymbol{\sigma}^2_{t-1}\{(\boldsymbol{a}_t,t)\}] \leq w(\sigma^{-2}) =c_4$, i.e.
\begin{eqnarray}\label{t1_a1_s3}
\sigma^{-2}\boldsymbol{\sigma}^2_{t-1}\{(\boldsymbol{a}_t,t)\} \leq  c_4\log\Big(1+ \sigma^{-2}\boldsymbol{\sigma}^2_{t-1}\{(\boldsymbol{a}_t,t)\} \Big).
\end{eqnarray} 

Based on inequality \ref{t1_a1_s3}, we have
\begin{eqnarray*}
\begin{split} 
R_T  \leq \sqrt{4 T {\beta_T}\sigma^{2} \sum_{t=1}^T \sigma^{-2}\boldsymbol{\sigma}^2_{t-1}\{(\boldsymbol{a}_t,t)\}}&\leq \sqrt{4 T {\beta_T}\sigma^{2} \sum_{t=1}^T c_4\log\Big(1+ \sigma^{-2}\boldsymbol{\sigma}^2_{t-1}\{(\boldsymbol{a}_t,t)\} \Big)}\\
&=  \sqrt{8 c_4 T {\beta_T}\sigma^{2} \Bigg\{{1\over2}\sum_{t=1}^T\log\Big(1+ \sigma^{-2}\boldsymbol{\sigma}^2_{t-1}\{(\boldsymbol{a}_t,t)\} \Big)\Bigg\}}.
\end{split}
\end{eqnarray*}  
Recall the established result in Lemma \ref{mig}, we have 
\begin{eqnarray}\label{t1_a1_bound_R_T}
R_T  \leq \sqrt{8 c_4 T {\beta_T}\sigma^{2}  I(\boldsymbol{y}_{1:T};\boldsymbol{f}_{1:T}) } \leq \sqrt{8 c_4 T {\beta_T}\sigma^{2}   \gamma^ \mathcal{S}_{T}}, 
\end{eqnarray}  
where $I(\boldsymbol{y}_{1:T};\boldsymbol{f}_{1:T})$ is bounded by the maximum mutual information gain $\gamma^ \mathcal{S}_{T}$. 

Lastly, let $c_3= 8 c_4 \sigma^{2} = 8/\log(1+ \sigma^{-2})$, we have
\begin{eqnarray*}
Pr \Bigg\{ R_T \leq \sqrt{ c_3 T  \beta_T\gamma^ \mathcal{S}_{T} } , \quad \forall T \geq 1\Bigg\} \geq 1 -\delta. \quad \blacksquare
\end{eqnarray*}

\subsection{Continuous Action Space}

Next, we consider a continuous action space $\mathcal{A} \subset [-v/2, v/2]^d$ that is compact and convex with $d \in \mathbb{N}$ and a constant $v > 0$. Suppose $f$ is sampled from a known GP prior with known noise variance $\sigma^2$. Give $\beta_t = 2 log\{  \pi^2 t^2 /(3\delta)\} +   2d \log\{c_2  v d t^2\sqrt{\log(c_1d \pi^2 t^2/(3\delta)}\}$, we show the regret bound of Periodic-GP-UCB by an additional discretization technique on the action space in following three steps.

\textbf{Step 1 (action discretization):}

First of all, take discretization $\mathcal{A}_t$ of size $\omega_t^d$ on the action space $\mathcal{A}$ at time step $t$ such that 
\begin{eqnarray*}
||\boldsymbol{a} -[\boldsymbol{a}]_t ||_1 \leq vd/\omega_t, \quad \forall \boldsymbol{a} \in \mathcal{A}_t,
\end{eqnarray*}
where $[\boldsymbol{a}]_t$ is the closest point to $\boldsymbol{a}$ in space $\mathcal{A}_t$.

Let $\omega_t = c_2 vdt^2\sqrt{\log \{ c_1d\pi^2 t^2/(3\delta)\}}$, then the size of $\mathcal{A}_t$ can be characterized by
 \begin{eqnarray*}\label{t1_a2_s1}
|\mathcal{A}_t| = \omega_t^d =  \Big[ c_2 vdt^2\sqrt{\log \{ c_1d\pi^2 t^2/(3\delta)\}}\Big]^d.
\end{eqnarray*}

\textbf{Step 2 (bound error):}

Recall $f$ satisfies the following high
probability bound on the derivatives of GP sample paths for some constants $c_1,c_2 > 0$ at time $t$ where $\boldsymbol{a}=[a^{(1)},\cdots,a^{(d)}]^\top$:
\begin{eqnarray*}
Pr \Bigg\{\sup_{\boldsymbol{a}\in\mathcal{A}} \Bigg|{\partial f(\boldsymbol{a},t) \over \partial a^{(j)}} \Bigg| \geq L_t \Bigg\} \leq c_1 \exp\{-(L_t/c_2)^2\},\quad j = 1,\cdots, d.
\end{eqnarray*} 

By union bound over $j = 1,\cdots, d$, we have 
\begin{eqnarray*}
Pr \Bigg\{ \Bigg|{\partial f(\boldsymbol{a},t) \over \partial a^{(j)}} \Bigg| < L_t \Bigg\} \leq 1-  c_1d \exp\{-(L_t/c_2)^2\},\quad \forall j, \quad \forall \boldsymbol{a}\in\mathcal{A}.
\end{eqnarray*} 

The above result leads to the Lipschitz continuity on $f$ such that 
$$
|f(\boldsymbol{a},t) - f(\boldsymbol{a}',t)| \leq L_t||\boldsymbol{a} -\boldsymbol{a}'||_1, \quad \forall \boldsymbol{a},\boldsymbol{a}' \in \mathcal{A}
$$
holds with probability at least $1-c_1d \exp\{-(L_t/c_2)^2\}$ for a time step $t$.

Let $L_t=c_2\sqrt{\log \{c_1d\pi^2 t^2/(3\delta)\}}$, then with definition of $\omega_t$, we have
\begin{eqnarray*}
L_t||\boldsymbol{a} -[\boldsymbol{a}]_t ||_1 \leq L_t vd/\omega_t = c_2 vd\sqrt{\log \{c_1d\pi^2 t^2/(3\delta)\}} /[c_2 vdt^2\sqrt{\log \{ c_1d\pi^2 t^2/(3\delta)\}}]=1/t^2.
\end{eqnarray*}

By taking union bound over $t\in \mathcal{T}$, we have 
\begin{eqnarray*}
|f(\boldsymbol{a},t) - f([\boldsymbol{a}]_t,t)| \leq L_t||\boldsymbol{a} -[\boldsymbol{a}]_t ||_1 \leq 1/t^2, \quad \forall \boldsymbol{a} \in \mathcal{A}_t,, \quad \forall t\geq 1,
\end{eqnarray*} 
holds with probability at least 
\begin{eqnarray}\label{res_2}
1-\sum_{t=1}^\infty c_1d \exp\{-(L_t/c_2)^2\} = 1-\sum_{t=1}^\infty c_1d \exp\{-(c_2\sqrt{\log \{c_1d\pi^2 t^2/(3\delta)\}}/c_2)^2\}  =1-\sum_{t=1}^\infty 3\delta/ (t^2 \pi^2) = 1- \delta/2. 
\end{eqnarray}

\textbf{Step 3 (bound regret):}

Similarly, fix a time step $t$ and an action $\boldsymbol{a} \in \mathcal{A}_t$, we have  
\begin{eqnarray*}
Pr\Big\{|f(\boldsymbol{a}, t) - \boldsymbol{\mu}_{t-1}\{(\boldsymbol{a},t)\} |> \sqrt{\beta_t} \boldsymbol{\sigma}_{t-1} \{(\boldsymbol{a},t)\} \Big\} \leq \exp\{-{1\over2}\beta_t\},
\end{eqnarray*}
where $\beta_t= 2 log\{  \pi^2 t^2 /(3\delta)\} +   2d \log\{c_2  vd t^2\sqrt{\log(c_1d \pi^2 t^2/(3\delta)}\}$.

By union bound over $\boldsymbol{a} \in \mathcal{A}_t$ and time step $t \in  \mathcal{T}$, we have 
\begin{eqnarray*}
 |f(\boldsymbol{a}, t) - \boldsymbol{\mu}_{t-1}\{(\boldsymbol{a},t)\} |\leq \sqrt{\beta_t} \boldsymbol{\sigma}_{t-1} \{(\boldsymbol{a},t)\} , \quad \forall \boldsymbol{a} \in \mathcal{A}_t, \quad \forall t \in  \mathcal{T},
 \end{eqnarray*}
holds with probability at least
\begin{eqnarray}\label{res_1}
\begin{split}
1 -  \sum_{t=1}^\infty |\mathcal{A}_t| \exp\{-{1\over2}\beta_t\} &= 1 - \sum_{t=1}^\infty \omega_t^d \exp\{-{1\over2}\beta_t\} \\
&=1 - \sum_{t=1}^\infty {\Big[c_2 vdt^2\sqrt{\log \{ c_1d\pi^2 t^2/(3\delta)\}}\Big]^d \over \exp\Big[ log\{  \pi^2 t^2 /(3\delta)\Big] \exp\Big[d \log\{c_2  vd t^2\sqrt{\log(c_1d \pi^2 t^2/(3\delta)}\}\Big] }\\
&= 1 - \sum_{t=1}^\infty 3\delta/ (t^2 \pi^2) = 1- \delta/2. 
\end{split}
 \end{eqnarray}

Based on the results in \ref{res_1} and \ref{res_2}, with probability at least $1- \delta$, we have
\begin{eqnarray*}
  f(\boldsymbol{a}_t^*, t)\leq f([\boldsymbol{a}_t^*]_t, t) + 1/t^2 \leq \boldsymbol{\mu}_{t-1}\{([\boldsymbol{a}_t^*]_t,t)\} + \sqrt{\beta_t} \boldsymbol{\sigma}_{t-1} \{([\boldsymbol{a}_t^*]_t,t)\} +1/t^2 .
 \end{eqnarray*}
 
Therefore, with probability at least $1- \delta$, the instantaneous regret incurred at time $t$ can be bound by
\begin{eqnarray}\label{t1_a1_bound_r_t}
\begin{split} 
r_t &=f(\boldsymbol{a}^*_t,t) -f(\boldsymbol{a}_t,t)\leq f([\boldsymbol{a}_t^*]_t, t) - f(\boldsymbol{a}_t,t) + 1/t^2 \\
&\leq  \boldsymbol{\mu}_{t-1}\{([\boldsymbol{a}_t^*]_t,t)\} + \sqrt{\beta_t} \boldsymbol{\sigma}_{t-1} \{([\boldsymbol{a}_t^*]_t,t)\}  - [\boldsymbol{\mu}_{t-1}\{(\boldsymbol{a}_t,t)\} - \sqrt{\beta_t} \boldsymbol{\sigma}_{t-1} \{(\boldsymbol{a}_t,t)\}] +1/t^2\\
&\leq \boldsymbol{\mu}_{t-1}\{(\boldsymbol{a}_t,t)\} + \sqrt{\beta_t} \boldsymbol{\sigma}_{t-1} \{(\boldsymbol{a}_t,t)\} - [\boldsymbol{\mu}_{t-1}\{(\boldsymbol{a}_t,t)\} - \sqrt{\beta_t} \boldsymbol{\sigma}_{t-1} \{(\boldsymbol{a}_t,t)\}]  +1/t^2 = 2 \sqrt{\beta_t} \boldsymbol{\sigma}_{t-1} \{(\boldsymbol{a}_t,t)\} + 1/t^2,\\
\end{split}
\end{eqnarray}

Following the same logic in Section \ref{discrete}, the cumulative regret up to time $T$ is bounded by
\begin{eqnarray*} 
R_T= \sum_{t=1}^T r_t \leq  \sum_{t=1}^T1/t^2 + \sqrt{T \sum_{t=1}^T 4  {\beta_t} \boldsymbol{\sigma}^2_{t-1}\{(\boldsymbol{a}_t,t)\}} \leq  \pi^2/6 + \sqrt{ c_3 T  \beta_T\gamma^ \mathcal{S}_{T}}, 
\end{eqnarray*}  
 with probability at least $1- \delta$, where $c_3= 8/\log(1+ \sigma^{-2})$.

Equivalently, we have
\begin{eqnarray*}
Pr \Bigg\{ R_T \leq \sqrt{ c_3 T  \beta_T\gamma^ \mathcal{S}_{T} } + \pi^2/6, \quad \forall T \geq 1\Bigg\} \geq 1 -\delta. \quad \blacksquare
\end{eqnarray*}

%
%
%
%

\section{Proof on Bounding $\gamma^ \mathcal{S}_{T}$ in Periodic-GP}
In this section, we provide details for the proof of Theorem 2, which characterize $\gamma^ \mathcal{S}_{T}$ in Periodic-GP. Let the circle $\tau$ is a fixed constant. Suppose $f$ is sampled from a known GP prior with known noise variance $\sigma^2$. We first bound the mutual information gain by rearranging the time steps $\{1,\cdots, T\}$ into $\tau$ sets of length $T/\tau$, such that within each set the function $f$ is time-invariant, by following proof of Lemma 1.

\subsection{Proof of Lemma 1}

The proof can be completed by the chain rule for the mutual information gain and the fact that the noise is independent, following Lemma 7.9.2 in \cite{cover1999elements}. 

First, recall the results in the proof of Lemma \ref{mig}, by the definition of the mutual information gain, we have 
$$
I(\boldsymbol{y}_{1:T};\boldsymbol{f}_{1:T}) = H(\boldsymbol{y}_{1:T}) - H(\boldsymbol{y}_{1:T}|\boldsymbol{f}_{1:T}). 
$$

Given the sequence of reward received $\boldsymbol{y}_{1:T} =[y_1,\cdots,y_T]^\top$, let $\boldsymbol{y}^{(i)} =[y_i, y_{i+\tau},\cdots,y_{i+(K-1)\tau}]^\top$, so we have $\boldsymbol{y}_{1:T} =\{\boldsymbol{y}^{(i)}\}_{1\leq i \leq \tau}$. From the fact that the entropy of a collection of random variables is less than the sum of their individual entropies, the first entropy term above can be bounded by
\begin{eqnarray}\label{t3_s1}
H(\boldsymbol{y}_{1:T}) \leq \sum_{i=1}^\tau H\{\boldsymbol{y}^{(i)}\}. 
\end{eqnarray}

On the other hand, based on the chain rule for the conditional entropy, we represent the second entropy term as
\begin{eqnarray*} 
H(\boldsymbol{y}_{1:T}|\boldsymbol{f}_{1:T}) =  \sum_{t=1}^T H\{y_t|\boldsymbol{y}_{1:(t-1)},\boldsymbol{f}_{1:t}\}.
\end{eqnarray*}

Since $y_t =f(\boldsymbol{a}_t,t) +  \epsilon_t$ with independent noise $\epsilon_t$, we have 
\begin{eqnarray*} 
H\{y_t|\boldsymbol{y}_{1:(t-1)},\boldsymbol{f}_{1:t}\}= H\{y_t| f(\boldsymbol{a}_t,t)\},
\end{eqnarray*}
which leads to
\begin{eqnarray}\label{t3_s2}
\begin{split} 
H(\boldsymbol{y}_{1:T}|\boldsymbol{f}_{1:T})  =  \sum_{t=1}^T H\{y_t|f(\boldsymbol{a}_t,t)\} &= \sum_{i=1}^\tau \sum_{k=1}^K H\{y_{i+(k-1)\tau}|f(\boldsymbol{a}_{i+(k-1)\tau},{i+(k-1)\tau})\} \\
&\geq \sum_{i=1}^\tau \sum_{k=1}^K H\{y_{i+(k-1)\tau}|\boldsymbol{f}^{(i)}\} \\&\geq  \sum_{i=1}^\tau H\{\boldsymbol{y}^{(i)}|\boldsymbol{f}^{(i)}\},
\end{split}
\end{eqnarray} 
where $\boldsymbol{f}^{(i)} =\{f(\boldsymbol{a}_j,i)\}_{j=i+(k-1)\tau, k=1,\cdots,K}$. Here, the first inequality comes from the property of conditional entropy, and the second inequality follows the fact that the entropy of a collection of random variables is less than the sum of their individual entropies.

Thus, based on the results in \ref{t3_s1} and \ref{t3_s2}, we have 
\begin{eqnarray}\label{bnd_mig}
\begin{split}
I(\boldsymbol{y}_{1:T};\boldsymbol{f}_{1:T}) &\leq \sum_{i=1}^\tau H\{\boldsymbol{y}^{(i)}\} - \sum_{i=1}^\tau H\{\boldsymbol{y}^{(i)}|\boldsymbol{f}^{(i)}\}\\
&= \sum_{i=1}^\tau \Big[H\{\boldsymbol{y}^{(i)}\} -  H\{\boldsymbol{y}^{(i)}|\boldsymbol{f}^{(i)}\} \Big]\\
&= \sum_{i=1}^\tau I\{\boldsymbol{y}^{(i)};\boldsymbol{f}^{(i)}\}. \quad  \blacksquare
\end{split}
\end{eqnarray}
   
   \subsection{Proof of Theorem 2}
   
Lastly, we show the bound for $\gamma^ \mathcal{S}_{T}$ based on the results in Lemma 1.

 Recall the maximum mutual information gain under a stationary $f_0$ up to time step $T$ for the action space:
\begin{eqnarray*}
\gamma^ \mathcal{A}_{T}:=\max_{A\subset \mathcal{A}:|A|=T} I(\boldsymbol{y}_{A};f_0) =\max_{A\subset \mathcal{A}:|A|=T} {1\over2}\log | \boldsymbol{I} + \sigma^{-2}\boldsymbol{K}_A |,
\end{eqnarray*}
where $A$ is a subset of  $\mathcal{A}$ with size $T$, $\boldsymbol{y}_{A} =\{y(\boldsymbol{a})\}_{\boldsymbol{a}\in \mathcal{A}}$, and $I(\boldsymbol{y}_{A};f_0) = H(\boldsymbol{y}_{A}) - H(\boldsymbol{y}_{A}|f_0)$ quantifies the reduction in uncertainty (measured in terms of difference of the entropy) about $f_0$ achieved by revealing $\boldsymbol{y}_{A}$, and $\boldsymbol{K}_A = [k_\mathcal{A}(\boldsymbol{a}, \boldsymbol{a}')]_{\boldsymbol{a}, \boldsymbol{a}'\in A}$ is the Gram matrix of $k_\mathcal{A}$ evaluated on set $A \subset \mathcal{A}$.

And for the periodic stationary model, the maximum mutual information gain depends on $\boldsymbol{y}_{S}= \{y(\boldsymbol{s})\}_{\boldsymbol{s}\in \mathcal{S}}$ of the joint action-time pairs $\boldsymbol{s} = (\boldsymbol{a},t)$ for $t=1,2,\cdots, T$ and $\boldsymbol{a}\in A\subset \mathcal{A}$ with $|A|=T$, and $f$:
\begin{eqnarray*} 
\gamma^ \mathcal{S}_{T}&:=\max_{A\subset \mathcal{A}:|A|=T} I(\boldsymbol{y}_{S};f) =  \max_{A\subset \mathcal{A}:|A|=T}  {1\over2}\log | \boldsymbol{I} + \sigma^{-2}\boldsymbol{K}_S |, 
\end{eqnarray*}
where the kernel matrix $\boldsymbol{K}_S  = [k(\boldsymbol{s}, \boldsymbol{s}')]_{\boldsymbol{s}, \boldsymbol{s}'\in S} = [k_\mathcal{A}(\boldsymbol{a},\boldsymbol{a}') \times k_\tau(t, t').
]_{(\boldsymbol{a},t), (\boldsymbol{a}',t')\in S} = \boldsymbol{K}_A \circ \boldsymbol{K}_\tau$. Here, note the information of time is given by the system and cannot be changed, so the second element of $\boldsymbol{s}$ is always fixed by its current time step. 

By maximizing both sides of inequality \ref{bnd_mig} over $\{\boldsymbol{a}_1,\cdots,\boldsymbol{a}_T\}$ (since the time $t$ is fixed at each step), we obtain the maximum mutual information $\gamma^ \mathcal{S}_{T}$ in Periodic-GP satisfies 
\begin{eqnarray*} 
\gamma^ \mathcal{S}_{T} \leq  \sum_{i=1}^\tau  \gamma^ \mathcal{A}_{T/\tau} = \tau  \gamma^ \mathcal{A}_{K}.\quad \blacksquare
\end{eqnarray*}

  \newpage
\section{Sensitivity Analysis}

In this section, we provide sensitivity analysis for the proposed Periodic-GP-UCB  when the cycle $\tau$ is misspecified. Typically, we repeat the real data applications conducted in Section 5 of the main text but allow the period $\tau$ chosen from $\{6, 12, 20,22,24,26,28\}$. The results of the cumulative regret are reported in the following Figure \ref{supp_real_res}, with comparison to the second and third best method in each dataset. In summary, Figure \ref{supp_real_res}, one can observe that our proposed Periodic-GP-UCB still maintains a good performance under minor assumption violation such as $\tau\in \{ 20,22, 26,28\}$ for the Madrid traffic pollution data.
\begin{figure}[!thp]
\centering
\begin{subfigure}{}
  \centering
\includegraphics[width=0.45\textwidth]{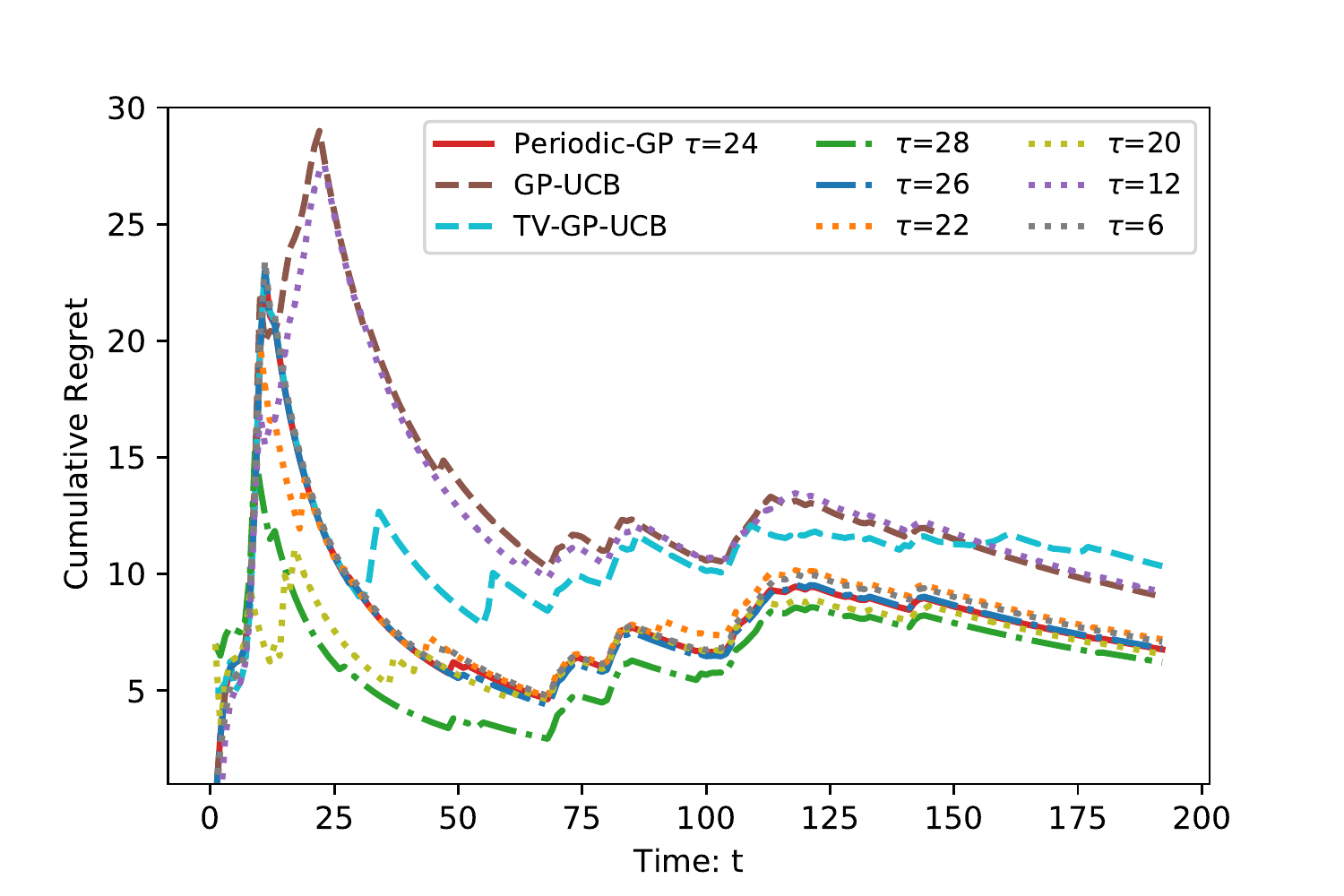} 
\end{subfigure} 
\caption{The cumulative regret under different methods for the Madrid traffic pollution data.} \label{supp_real_res}
\end{figure}

\end{document}